\documentclass[letterpaper]{article} 
\usepackage{aaai25}  
\usepackage{times}  
\usepackage{helvet}  
\usepackage{courier}  
\usepackage[hyphens]{url}  
\usepackage{graphicx} 
\usepackage{natbib}  
\usepackage{caption} 
\usepackage{algorithm}
\usepackage{algorithmic}
\usepackage{amsmath} 
\usepackage{amssymb}
\usepackage{multirow} 
\usepackage{pdfpages}

\usepackage{newfloat}
\usepackage{listings}
\DeclareCaptionStyle{ruled}{labelfont=normalfont,labelsep=colon,strut=off} 
\floatstyle{ruled}
\newfloat{listing}{tb}{lst}{}
\floatname{listing}{Listing}
\title{Anti-Diffusion: Preventing Abuse of Modifications of Diffusion-Based Models}
\author{
    Li Zheng\textsuperscript{\rm 1}\equalcontrib,
    Liangbin Xie\textsuperscript{\rm 1 \rm 2}\equalcontrib,
    Jiantao Zhou\textsuperscript{\rm 1}\thanks{Corresponding author},
    Xintao Wang\textsuperscript{\rm 3},
    Haiwei Wu\textsuperscript{\rm 1},
    Jinyu Tian\textsuperscript{\rm 4}
}
\affiliations{
    \textsuperscript{\rm 1}University of Macau\\
    \textsuperscript{\rm 2}Shenzhen Institute of Advanced Technology \\
    \textsuperscript{\rm 3}Kuaishou Technology\\
     \textsuperscript{\rm 4}Macau University of Science and Technology

    yc27908@um.edu.mo,
    lb.xie@siat.ac.cn,
    jtzhou@um.edu.mo,\\
    xintao.alpha@gmail.com,
    yc07912@um.edu.mo,
    jytian@must.edu.mo
}

\usepackage{bibentry}

\begin{document}

\maketitle

\begin{abstract}
Although diffusion-based techniques have shown remarkable success in image generation and editing tasks, their abuse can lead to severe negative social impacts.
Recently, some works have been proposed to provide defense against the abuse of diffusion-based methods.
However, their protection may be limited in specific scenarios by manually defined prompts or the stable diffusion (SD) version.
Furthermore, these methods solely focus on tuning methods, overlooking editing methods that could also pose a significant threat. 
In this work, we propose Anti-Diffusion, a privacy protection system designed for general diffusion-based methods, applicable to both tuning and editing techniques.
To mitigate the limitations of manually defined prompts on defense performance, we introduce the prompt tuning (PT) strategy that enables precise expression of original images. 
To provide defense against both tuning and editing methods, we propose the semantic disturbance loss (SDL) to disrupt the semantic information of protected images.
Given the limited research on the defense against editing methods, we develop a dataset named Defense-Edit to assess the defense performance of various methods.
Experiments demonstrate that our Anti-Diffusion achieves superior defense performance across a wide range of diffusion-based techniques in different scenarios.
\end{abstract}

%
\begin{links}
     \link{Code}{https://github.com/whulizheng/Anti-Diffusion}
\end{links}

\section{Introduction}
\label{sec:intro}

\begin{figure}[ht]
\centering
\includegraphics[width=0.7\columnwidth]{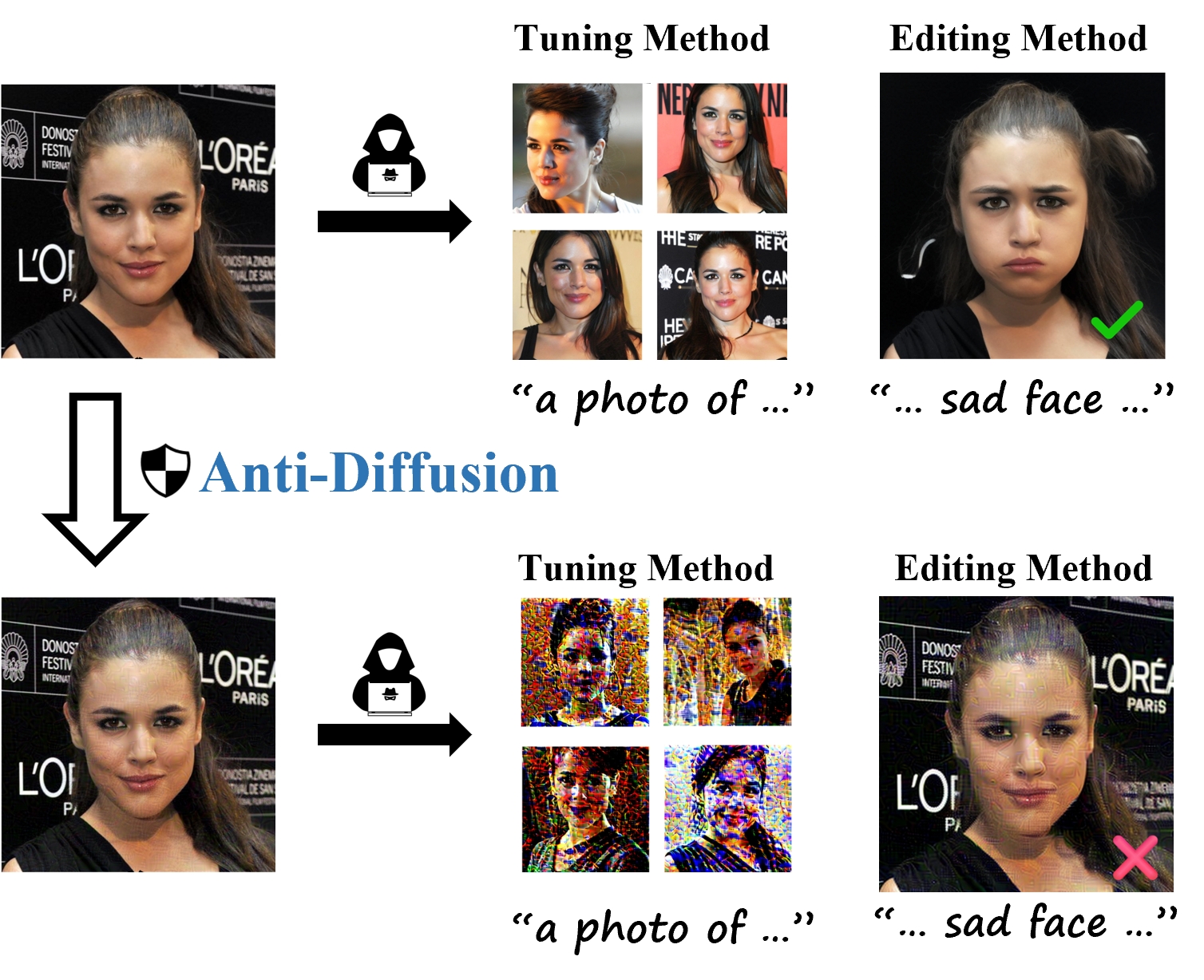}
\centering
\caption{Our defense system, called Anti-Diffusion, can provide defense against both tuning and editing methods.}
\label{fig:teasor} 
\end{figure}

The field of text-to-image synthesis~\cite{li2023gligen,ramesh2021zero,gafni2022make,ding2021cogview} has experienced significant advancements, primarily driven by diffusion models~\cite{ho2020denoising,song2020denoising}. 
Numerous diffusion models have demonstrated their ability to generate images of exceptional quality, such as SD ~\cite{rombach2022high,yang2023diffusion}, Pixel-Art~\cite{chen2023pixartalpha,chen2024pixartdelta}.
Based on these diffusion models, some controllable generation methods (ControlNet~\cite{zhang2023adding}, T2I-Adapter~\cite{mou2023t2i}) and personalized methods (DreamBooth~\cite{ruiz2023dreambooth}, LoRA~\cite{hu2021lora}, Textual Inversion~\cite{gal2022image}) have also been proposed.
With the rapid advancement of text-to-image techniques, many industry professionals and even ordinary users could create images or train personalized models based on their ideas.

However, technology is a double-edged sword. 
Individuals can easily utilize images to train personalized models (e.g., DreamBooth, LoRA) and manipulate images using editing methods such as MasaCtrl~\cite{cao_2023_masactrl} and DiffEdit~\cite{couairon2023diffedit}. 
Similar to DeepFake~\cite{liu2023deepfacelab,rana2022deepfake}, when these methods are abused by malicious users to create fake news, plagiarize artistic creations, violate personal privacy, etc., they can have severe negative impacts on both individuals and society~\cite{wang2023security}.
Hence, finding ways to protect images from the potential abuse of these methods is a pressing issue that requires immediate attention.

Anti-DreamBooth (Anti-DB)~\cite{van2023anti} has made attempts to address this issue. 
By adding subtle adversarial noise to images, Anti-DB forces the personalized model trained on them producing outputs with significant visual artifacts.
However, Anti-DB demands additional substitute data and manually defined prompts, which increases its complexity of use. Moreover, in practical scenarios, it is challenging to anticipate the prompts that malicious users might utilize, thereby limiting its defense performance.
Additionally, existing methods~\cite{truong2024attacks} focus solely on defending against personalized generative models, overlooking another crucial scenario—defense against editing models. Editing models have the capability to directly modify the content of input images during inference using prompts, thereby presenting a significant security and privacy threat if abused.

In this work, we propose Anti-Diffusion, a privacy protection system to prevent images from being abused by general diffusion-based methods. 
This system aims to add subtle adversarial noise~\cite{goodfellow2014explaining} to users' images before publishing in order to disrupt the tuning and editing process of diffusion-based methods.
To mitigate the impact of different prompts when defending and malicious using, and to overcome limitations of manually defined prompts in achieving optimal performance, as shown in Tab.~\ref{tab:prompt}, we propose the prompt tuning (PT) strategy. This strategy aims to optimize a text embedding that more accurately captures the information of protected images.
Our method with PT does not require manual selection of prompts during the defense phase and still provides good protection against malicious users training with unknown prompts.
Furthermore, as SD achieves semantic control of images through cross-attention~\cite{vaswani2017attention}, we introduce the semantic disturbance loss (SDL) to disrupt the semantic information of protected images. By minimizing the distance between the cross-attention map and a zero-filled map, it can maximize the semantic distance between clean images and protected images. When equipped with these two designs, our Anti-Diffusion can achieve robust defense against both tuning and editing methods, as shown in Fig.~\ref{fig:teasor}. To better evaluate the effectiveness of current defense methods against diffusion-based editing methods, in this work, we further construct a dataset, named Defense-Edit. We hope this dataset can draw attention to the privacy protection challenges posed by diffusion-based image editing models.
In summary, our contributions are as follows: 

\textbf{1)} We expand the defense to include both tuning-based and editing-based methods, while other baselines focus only on tuning-based methods. 

\textbf{2)} We introduce the PT strategy for ensuring a better representation of protected images and providing more generalized protection for unexpected prompts.

\textbf{3)} We integrate the SDL to disrupt the semantic information of protected images, enhancing the performance of defense against both tuning-based and editing-based methods.

\textbf{4)} We contribute a dataset called Defense-Edit for evaluating the defense performance against editing-based methods.

Based on both quantitative and qualitative results, our proposed method, Anti-Diffusion, achieves superior defense effects across several diffusion-based techniques, including tuning methods (such as DreamBooth/LoRA) and editing methods (such as MasaCtrl/DiffEdit). 

\begin{table}[ht]
\begin{center}

\begin{tabular}{c|c|ccc}
\hline
Defense          & Test                      & FDFR↑         & ISM↓                & BRISQUE↑       \\ \hline
Anti-DB(c1) & \multirow{3}{*}{c1} & {\underline{0.60}}    & {\underline{ 0.24}}        & {\underline{ 37.41}}    \\
Anti-DB(c2) &                         & 0.48          & 0.20                    & 37.21          \\
Anti-Diffusion        &                         & \textbf{0.62} & \textbf{0.15}  & \textbf{40.46} \\ \hline
Anti-DB(c1) & \multirow{3}{*}{c2} & 0.37          & 0.27                  & 36.37          \\
Anti-DB(c2) &                         & {\underline{ 0.40}}    & {\underline{ 0.25}}        & {\underline{ 36.96}}    \\
Anti-Diffusion        &                         & \textbf{0.60} & \textbf{0.17}  & \textbf{40.66} \\ \hline
\end{tabular}
\caption{Defense performance on the DreamBooth model with different prompts. c1 (``a photo of sks person''), c2 (``a dlsr portrait of sks person'').}
\label{tab:prompt}
\end{center}
\end{table}

\begin{figure*}[ht]
\centering
\includegraphics[width=0.85\linewidth]{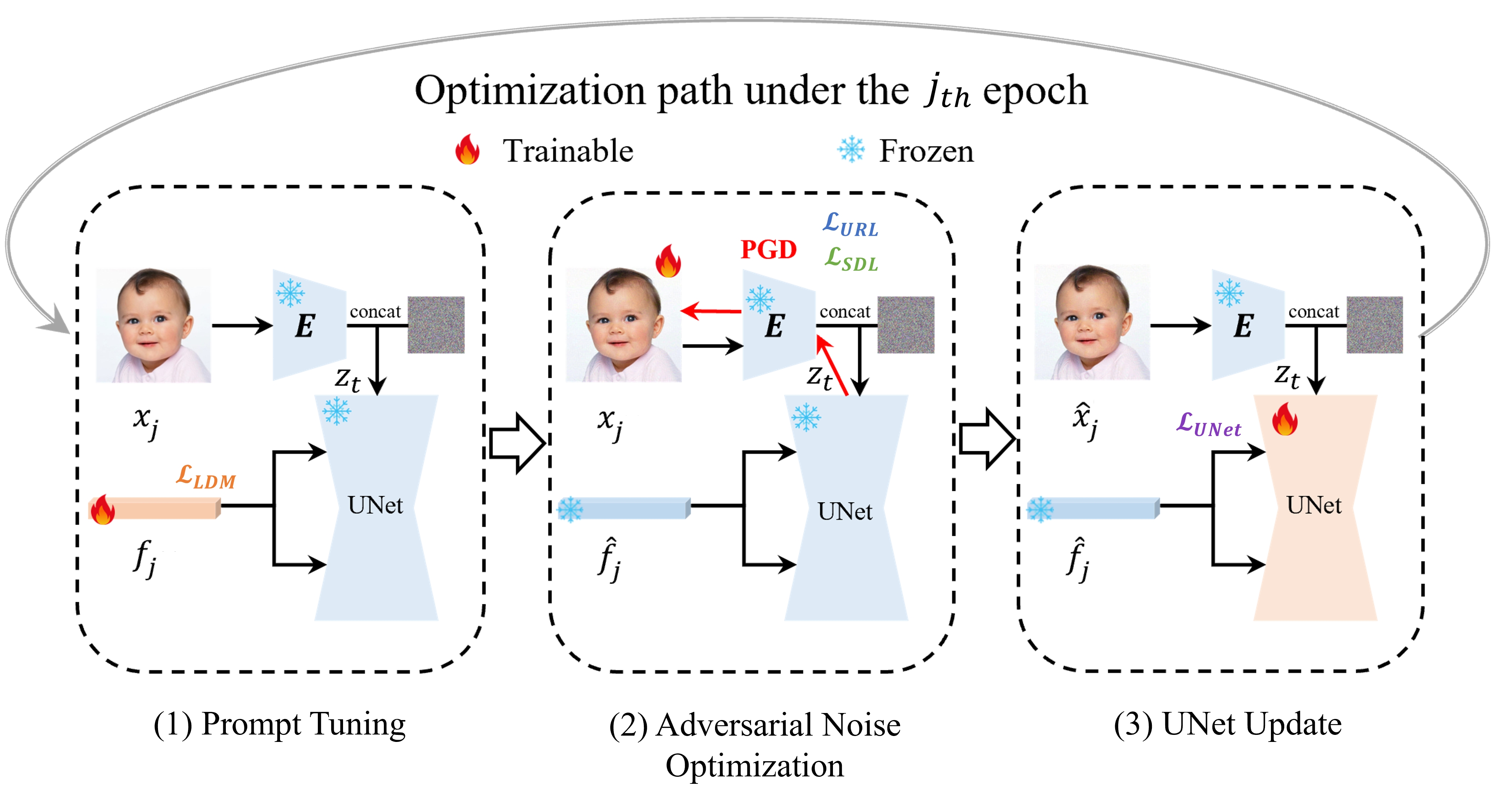}
\caption{The overview framework of Anti-Diffusion under the $j_{th}$ epoch. Here $x_{j}$ represents the image to be protected. In stage (1), the text-embedding $f_{j}$ will undergo fine-tuning with the $\mathcal{L}_{\mathrm{LDM}}$. Subsequently, in stage (2), adversarial noise will be optimized and added to $x_{j}$ using the PGD with our proposed loss functions $\mathcal{L}_{\mathrm{U R L}}$ and $\mathcal{L}_{\mathrm{S D L}}$ to obtain the adversarial sample $\hat{x}_{j}$. In stage (3), the UNet will be updated with $\mathcal{L}_{\mathrm{UNet}}$ using the adversarial sample $\hat{x}_{j}$ and text embedding $\hat{f}_{j}$ to simulate the tuning process of malicious users. This process repeats cyclically, returning to stage (1) in the next epoch.}
\label{fig:overview} 
\end{figure*}

\section{Preliminary}

\subsection{Stable Diffusion}
Stable diffusion is a Latent Diffusion Model (LDM) that has been trained on large-scale data. The LDM is a generative model capable of synthesizing high-quality images from Gaussian noise. Unlike traditional diffusion models, the diffusion process in LDM occurs in the latent space. Consequently, in addition to a diffusion model, an autoencoder, comprising an encoder $\mathcal{E}$ and a decoder $\mathcal{D}$, is required. For an image $x$ and an encoder $\mathcal{E}$, the diffusion process introduces noise to the encoded latent variable $z=\mathcal{E}(x)$, resulting in a noisy latent variable $z_{t}$, with the noise level escalating over timesteps $t \in T$. Subsequently, a UNet $\epsilon_\theta $ is trained to predict the noise added to the noisy latent variable $z_{t}$, given the text embedding instruction $f$. The specific loss function of latent diffusion is as follows:

\begin{equation}
\mathcal{L}_{l d m}:=\mathbb{E}_{z \sim \mathcal{E}(x), f, \epsilon \sim \mathcal{N}(0,1), t}\left[\left\|\epsilon-\epsilon_\theta\left(z_t, t, f\right)\right\|_2^2\right]
\label{equ:ldm}
\end{equation}

\subsection{Cross Attention Mechanism}
Attention mechanism allows models to refer to another related sequence when processing one sequence. It is an important part of diffusion, which introduces conditional information into the denoising process, thereby indicating the generated image. Many editing methods, such as MasaCtrl and DiffEdit, also use attention mechanisms to edit images. Cross-attention in diffusion can be expressed as:

\begin{equation}
    Attention(Q,K,V) = softmax(\frac{QK^T}{\sqrt{d}})\cdot V
    \label{equa-attention}
\end{equation}

\noindent where $Q=W_Q\cdot\varphi(z_t)$, $K=W_K\cdot f$ and $V=W_v\cdot f$. Here $\varphi(z_t)$ denotes a representation of the UNet implementing $\epsilon_\theta$, $d$ is used to ensure the normalization input of the softmax layer, and $W$ represents a learnable weight matrix.

\section{Methods} 

In this work, we aim to protect images by adding adversarial noise. 
We first provide a detailed definition of this problem.
Subsequently, we introduce the overall framework of Anti-Diffusion, which primarily encompasses three stages of iterative optimization. 
The first stage involves PT, and the second stage focuses on the optimization of adversarial noise, resulting in adversarial samples. The final stage involves updating the UNet with these adversarial samples.

\subsection{Problem Definition}

Recalling that our aim is to prevent the malicious use of diffusion-based image generation models on private images, 
we achieve this by adding adversarial noise to those images.
This adversarial noise disrupts the functionality of the malicious models while minimizing the visual impact on the images.
Let $x$ represent the image that requires protection. An adversarial noise $\delta$ is added, resulting in a protected image $\hat{x} = x+\delta$. 
The optimization of this adversarial noise $\delta$ can be described as a min-max optimization problem. 
The minimization simulates the actions of malicious users attempting to overcome the adversarial noise added to the protected images. 
The maximization aims to degrade the performance of the malicious model by adding adversarial noise under the constraint of maximal perturbations of the protected images. 
This min-max problem $P.1$ can be described as:

\begin{equation} \label{main}
\begin{array}{ll}
\centering
P.1:
 &  \min\limits_{\theta} ~ \max\limits_{\delta}~ \mathcal{L} \left(\epsilon_{\theta}, \hat {x},f\right) + \mathcal{C} \left(\epsilon_{\theta}, \hat {x}, {f} \right),\\
&\text { s.t. } \left\|\delta\right\|_p \leq \eta,
\end{array}
\end{equation}

\noindent where $\eta$ controls the $L_p$ norm perturbation magnitude of the adversarial noise $\delta$. $\mathcal{L}$ is the loss function of this generation model trained on the modified images.  $\mathcal{C}$ measures the feature dissimilarity of the images generated by the diffusion-based generation model $\epsilon_{\theta}$, the input image, and the target prompt. $f$ is the text embedding of the input prompt.
We generate the adversarial noise by maximizing the objective function $P.1$.
Then we optimize the model $\epsilon_{\theta}$ to minimize this function following the original training process of SD.




\subsection{Overview Framework}

To solve the min-max problem $P.1$, we need to apply alternating optimization over multiple epochs.
In each epoch, we divide this optimization into three stages: (1) prompt tuning, (2) adversarial noise optimization, and (3) UNet update, as illustrated in Fig.~\ref{fig:overview}. 
Specifically, stage 2 corresponds to the maximization of $P.1$ while stage 3 corresponds to the minimization of $P.1$. Given that an accurate text-embedding $f$ is crucial for $P.1$, we include stage 1 to train the text-embedding $f$ at the beginning of each epoch.

Fig.~\ref{fig:overview} illustrates the optimization path under the $j_{th}$ epoch. 
In the $j_{th}$ epoch, $x_{j}$ and $f_{j}$ are first input into stage prompt tuning. At this point, the parameters of the image encoder and UNet are fixed. We only optimize $f_{j}$ to obtain a better $\hat{f}_{j}$ that corresponds to the semantic information of the input image.
Subsequently, $x_{j}$ and the optimized $\hat{f}_{j}$ are incorporated into the adversarial noise optimization stage. 
In this stage, ${x}_{j}$ is continually optimized by utilizing the PGD algorithm with loss functions $\mathcal{L}_{\mathrm{U R L}}$ and $\mathcal{L}_{\mathrm{S D L}}$.
The adversarial sample $\hat{x}_{j}$ and $\hat{f}_{j}$ are input into the next UNet update stage to facilitate the update of the UNet parameters.
After the $j_{th}$ epoch, the updated $\hat{x}_{j}$, $\hat{f}_{j}$  and $\hat{\theta}$ will serve as the $x_{j+1}$, $f_{j+1}$ and $\theta_{j+1}$
Note that in the first stage, the image $x_0$ is initialized with a clean image, and the text embedding $f_0$ is the embedding of an empty prompt. After $N$ epochs, we obtain the final protected image $x_N$.

\subsection{Prompt Tuning Strategy}
\label{sec:prompt_tuning}

Due to the inability to predict what prompts malicious users will utilize to train their models, it is challenging for Anti-DB to manually define a prompt that can provide the best protection on different metrics.
Therefore, we propose the PT strategy to address this issue. 
As shown in Fig.~\ref{fig:overview} (1), we iteratively optimize $f_{j}$ under each epoch to obtain a more accurate representation corresponding to $x_{j}$. Initially, the image $x_{j}$ undergoes processing through the image encoder before being combined with the noise map to generate the noisy latent $z_{t}$. This noisy latent is then fed into the UNet, where it interacts with $f_{j}$ via cross-attention. We optimize $f_{j}$ to obtain $\hat{f}_{j}$ by using the loss function $\mathcal{L}_{\mathrm{LDM}}$ of the latent diffusion model. The parameters of the image encoder and UNet are fixed.
By continuously optimizing the text embedding $f$, the model can predict the correct noise; the semantics of $\hat{f}$ are expected to gradually align with the feature content of the images.

\subsection{Adversarial Noise Optimization}
Following the maximization of the function $P.1$ in the Problem Definition, we employ the projected gradient descent (PGD) algorithm~\cite{madry2018towards} to optimize the adversarial noise. 
The PGD algorithm is chosen for its convenience and efficiency.
We introduce   $\mathcal{L}_{\mathrm{URL}}$ and $\mathcal{L}_{\mathrm{SDL}}$ as the loss functions of PGD to interfere with the training process of SD and disturb the semantic information of protected images.
\subsubsection{PGD Optimization}
The PGD algorithm is used to optimize the adversarial noise added to images. With the two designed loss functions $\mathcal{L}_{\mathrm{URL}}$ and $\mathcal{L}_{\mathrm{SDL}}$, the cost function is as follows:
\begin{small}
\begin{equation}
\mathcal{C}=\mathcal{L}_{\mathrm{URL}}\left(x,\hat{f}_{j},\epsilon_\theta,\mathcal{E}\right) +  \mathcal{L}_{\mathrm{SDL}}\left(x,\hat{f}_{j},M_{target},\epsilon_\theta,\mathcal{E}\right),
\end{equation}
\end{small}

Using $p$ to represent the number of iterations of the current PGD, the gradient based on the cost equation $\mathcal{C}$ for the current $x_p$ can be calculated as:

\begin{equation}
    g_p=\nabla_{x_{p}}\mathcal{C}\left(x_{p},\hat{f}_{j},M_{target},\epsilon_\theta,\mathcal{E}\right),
\end{equation}

Therefore, the updated image $x_{p+1}$ with adversarial noise can be calculated as follows:

\begin{equation}
    x_{p+1}=\prod_{S}(x_{p}-|\alpha| \cdot sign(g_p)),
\end{equation}

\noindent where $ S=\{x_{p}|D(x_{p},x_{p+1})\leq\epsilon\} $ and  $\alpha$ represents the step size.
After all the iterations of the PGD attack, the adversarial samples ${\hat{x}_{j}}$ are updated from the clean images $x$ or the adversarial samples ${\hat{x}_{j-1}}$ from the previous epoch.

\subsubsection{UNet Reverse Loss}
Diffusion models generate or edit images by predicting noise from $z_t$, or learn the distribution of targets by predicting the added noise $\epsilon$ from the sampled $z_t$. To interfere with the prediction of noise by model $\epsilon_\theta$, the UNet Reverse Loss is designed as follows:
\begin{small}
\begin{equation}
\mathcal{L}_{\mathrm{URL}}:=\mathbb{E}_{z \sim \mathcal{E}(x), \hat{f}_{j}, \epsilon \sim \mathcal{N}(0,1), t}\left[-\left\|\epsilon-\epsilon_\theta\left(t, z_t, \hat{f}_{j}\right)\right\|_2^2\right]
\label{equ:url}
\end{equation}
\end{small}

\subsubsection{Semantic Disturbance Loss}
\begin{figure}[ht]
\centering
\includegraphics[width=0.7\linewidth]{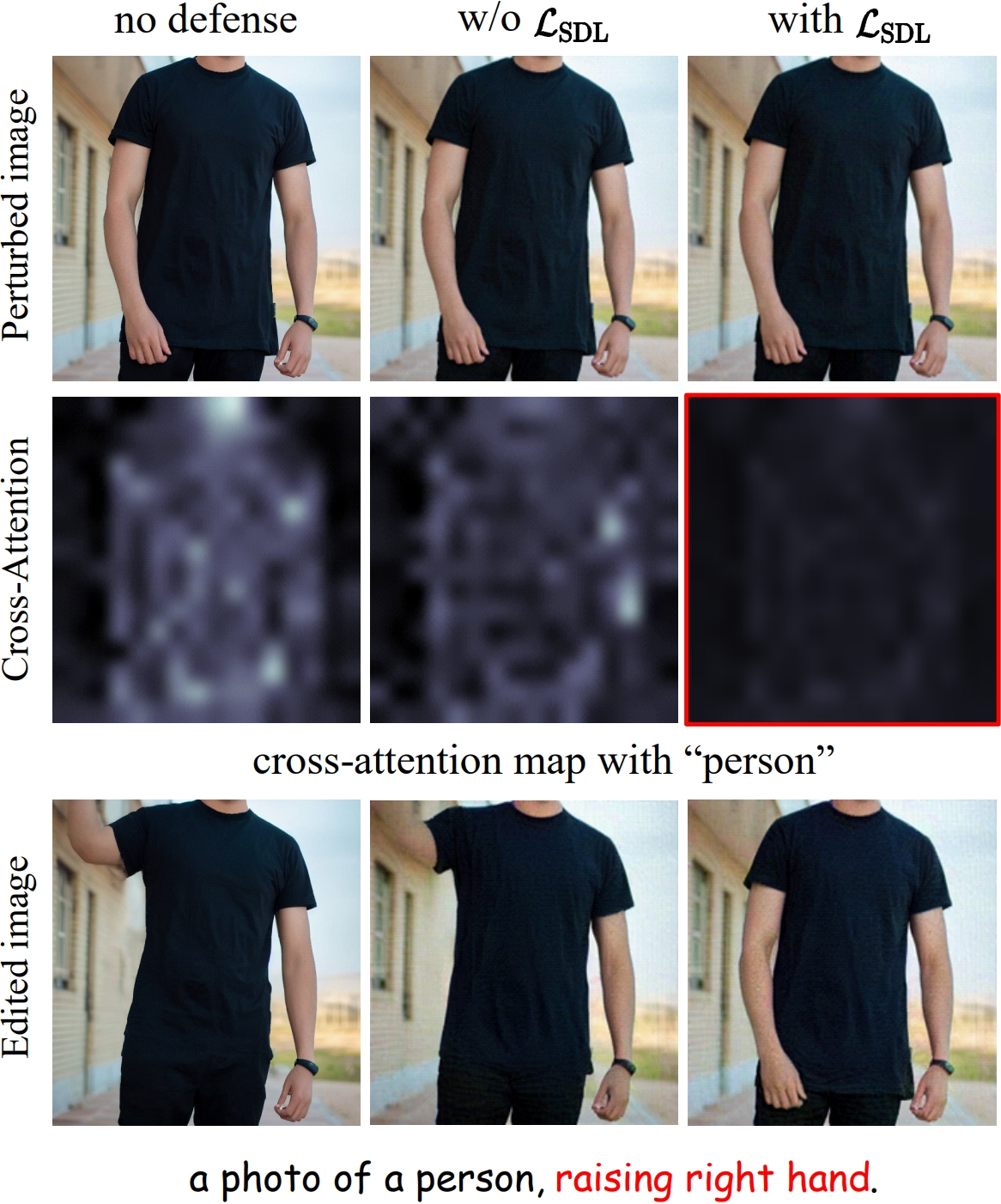}
\caption{Visualization results of how $\mathcal{L}_{\mathrm{SDL}}$ works. The editing method is DiffEdit.
}
\label{fig:ablation_sdl} 
\end{figure}

As depicted in Fig.~\ref{fig:ablation_sdl}, the cross-attention map represents the similarity between the relevant areas of the image and token. We design the SDL to interfere with the original semantic information of the protected image, rendering the editing method ineffective on the protected image. 
The $\mathcal{L}_{\mathrm{SDL}}$ is designed as follows:
\begin{scriptsize}
\begin{equation}
\mathcal{L}_{\mathrm{SDL}}:=\mathbb{E}_{z \sim \mathcal{E}(x), \hat{f}_{j}, \epsilon \sim \mathcal{N}(0,1), t}\left[\left\|M_{\mathrm{target}}-M\left(\epsilon_\theta,  t,  z_t,\hat{f}_{j}\right)\right\|_2^2\right],
\label{equ:sdl}
\end{equation}
\end{scriptsize}

\noindent where $M_{\mathrm{target}}$ is the target Attention map. In our experiments, we set it as a zero matrix with the same size as $M$. 

\subsection{UNet Update}
\label{sec:update_unet} 
Following the minimization of the function $P.1$ in the Problem Definition, we optimize the UNet model to simulate the behavior of malicious users for training their tuning-based methods.
This optimization is conducted with $\mathcal{L}_{\mathrm{UNet}}$ to further improve the defense performance of the proposed method against these tuning-based methods. 
Similar to the loss function of $\mathcal{L}_{\mathrm{LDM}}$, we optimize the UNet $\epsilon_\theta$ with adversarial sample ${\hat{x}_{j}}$   and  text embedding $\hat{f}_{j}$ based on the loss:
\begin{small}
\begin{equation}
\mathcal{L}_{\mathrm{UNet}}:=\mathbb{E}_{z \sim \mathcal{E}({\hat{x}_{j}}), \hat{f}_{j}, \epsilon \sim \mathcal{N}(0,1), t}\left[\left\|\epsilon-\epsilon_\theta\left(t, z_t, \hat{f}_{j}\right)\right\|_2^2\right]
\label{equ:unet}
\end{equation}
\end{small}

\section{Benchmark for Editing Methods}
\label{section:benchmark}

Existing research on defense against diffusion models primarily concentrates on personalized diffusion models like DreamBooth and LoRA, overlooking diffusion-based image editing methods such as MasaCtrl and DiffEdit. 
Diffusion-based editing methods, commonly used within the community, raise privacy protection concerns similar to personalized tuning models. Therefore, we have collected a dataset named Defense-Edit to additionally evaluate the defense performance against diffusion-based editing methods. The Defense-Edit dataset comprises a total of 50 pairs of images and prompts, including 30 pairs collected from CelebA-HQ, VGGFace2, TEdBench~\cite{kawar2023imagic}, and 20 pairs generated from SD. 
Additional details about Defense-Edit can be found in the supplementary materials.

\begin{table*}[ht]
\centering
\begin{tabular}{c|c|c|cccccc}
\hline
Dataset                    & Method          & PSNR↑          & FDFR↑         & ISM↓          & SER-FQA↓      & BRISQUE↑       & FID↑            & NIQE↑         \\ \hline
\multirow{6}{*}{VGGFace2}  & no defense      & —              & 0.10           & 0.66          & 0.73          & 17.43          & 144.02          & 4.12          \\
                            & MIST            & 34.35          & 0.03          & 0.60           & 0.85          & 26.46          & 204.35          & 4.51          \\
                           & Photo Guard     & 34.40          & 0.01          & 0.62          & 0.67          & 27.58          & 181.53          & 4.32          \\
                          & PID             & 34.62          & 0.42          & 0.51          & 0.53          & 32.57          & 301.53          & 4.75          \\
                          & Anti-DB & 34.55          & 0.60           & 0.24          & 0.31          & 37.41          & 436.34          & 5.05          \\
                          & Anti-Diffusion            & \textbf{35.91} & \textbf{0.62} & \textbf{0.15} & \textbf{0.18} & \textbf{40.46} & \textbf{457.13} & \textbf{5.27} \\ \hline
\multirow{6}{*}{CelebA-HQ} & no defense      & —              & 0.07          & 0.63          & 0.73          & 17.00             & 147.82          & 4.72          \\
                           & MIST            & 35.73          & 0.01          & 0.58          & 0.72          & 32.75          & 258.54          & 4.74          \\
                          & photo guard     & 35.35          & 0.08          & 0.49          & 0.69          & 24.34          & 217.58          & 4.68          \\
                         & PID             & 35.24          & 0.24          & 0.42          & 0.52          & 35.25          & 286.65          & 4.78          \\
                          & Anti-DB & 35.76          & 0.54          & 0.41          & 0.39          & 38.34          & 336.12          & 5.56          \\
                           & Anti-Diffusion            & \textbf{36.76} & \textbf{0.58} & \textbf{0.26} & \textbf{0.38} & \textbf{40.93} & \textbf{352.83} & \textbf{5.96} \\ \hline
\end{tabular}
\caption{Comparing the defense performance of different methods on the DreamBooth model. The inference prompt adopted in DreamBooth is ``a photo of sks person''. The best-performing defense under each metric is marked with \textbf{bold}.}
\label{tab:dreambooth_cmp}
\end{table*}

\begin{table*}[ht]
\centering
\begin{tabular}{ c|c|cccccc}
\hline
Method          & PSNR↑          & FDFR↑         & ISM avg↓      & SER-FQA↓      & BRISQUE↑      & FID↑            & NIQE↑         \\ \hline
no defense      & —              & 0.06          & 0.54          & 0.74          & 17.15          & 201.00          & 4.12          \\ \hline
Photo Guard     & 34.40          & 0.06          & 0.47          & 0.70          & 17.53          & 233.64          & 4.78          \\
MIST            & 34.35          & 0.07          & 0.43          & 0.58          & 16.24          & 256.26          & 4.95          \\
PID             & 34.62          & 0.15          & 0.46          & 0.61          & 20.62          & 295.15          & 5.42          \\
Anti-DB & 34.55          & \textbf{0.21} & 0.37          & 0.46          & 37.47          & 319.75          & 6.85          \\
Anti-Diffusion  & \textbf{35.91} & \textbf{0.21} & \textbf{0.35} & \textbf{0.45} & \textbf{39.26} & \textbf{326.28} & \textbf{7.18} \\ \hline
\end{tabular}
\caption{Comparing the defense performance of different methods on LoRA model on VGGFace2. The inference prompt adopted in LoRA is ``a photo of sks person''.}
\label{tab:lora_cmp}
\end{table*}

\section{Experiment}

\subsection{Implementation Details}

\subsubsection{Datasets.} To train the DreamBooth/LoRA models, we follow the dataset usage of the Anti-DB. Specifically, we conduct experiments using the 100 unique identifiers (IDs) gathered from VGGFace2~\cite{cao2018vggface2}  and CelebA-HQ~\cite{karras2017progressive} datasets. For the MasaCtrl/DiffEdit methods, we execute experiments based on our own collected Defense-Edit dataset.

\subsubsection{Evaluation Metrics.} To measure the defense performance on the DreamBooth and LoRA models, following Anti-DB, we also adopt these four metrics: BRISQUE~\cite{mittal2012no}, SER-FQA~\cite{terhorst2020ser}, FDFR~\cite{deng2020retinaface}, and ISM~\cite{deng2019arcface}.
We further introduce two additional IQA metrics, Fréchet Inception Distance (FID)~\cite{heusel2017gans} and Natural Image Quality Evaluator (NIQE)~\cite{mittal2012making}.
In addition, to measure the degradation of the visual quality of the original image caused by the addition of adversarial noise, we employ the Peak Signal-to-Noise Ratio (PSNR) metric~\cite{korhonen2012peak}. 
The CLIP Score measures the degree of alignment between a specific image and its target textual description. 
In our evaluation for editing methods like MasaCtrl and DiffEdit, the CLIP Score~\cite{hessel2021clipscore} is calculated by the edited images and target prompts. BRISQUE is also used to measure the image quality of edited images.
In our experiments, we aim for a lower CLIP Score and a higher BRISQUE Score.

\subsection{Comparison with State-of-the-art Methods}

\begin{figure*}[ht]
\centering
\includegraphics[width=0.8\linewidth]{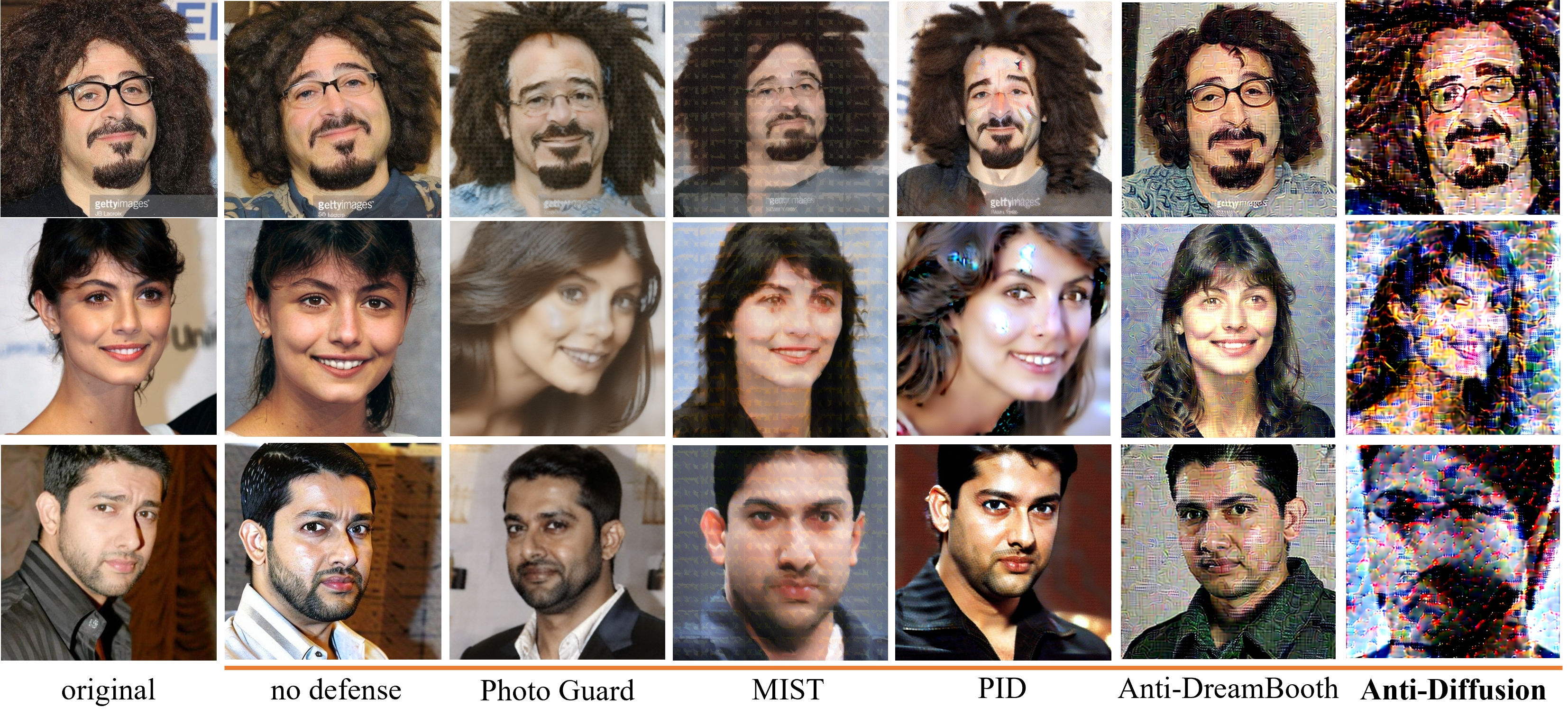}
\caption{Qualitative defense results of different methods on the DreamBooth model. 
The specific prompt adopted in DreamBooth is ``a photo of sks person''. The instance is from VGGFace2.
}
\label{fig:dreambooth_cmp} 
\end{figure*}

We compare Anti-Diffusion with state-of-the-art defense methods, namely Photo Guard~\cite{salman2023raising}, MIST~\cite{liang2023adv}, Anti-DB, and PID~\cite{li2024pid}. To ensure a fair comparison, following Anti-DB, we adopt the noise budget of $\eta=0.05$ for all these methods. 
During the evaluation process, for each trained DreamBooth/LoRA model, we generate $16$ images under $5$ different seeds, totaling $80$ images, to evaluate the corresponding results, thereby eliminating the variability associated with a single seed. 
For non-trainable diffusion-based image editing methods like MasaCtrl/DiffEdit, we evaluate the defense performance based on our Defense-Edit dataset.

\subsubsection{Comparison on DreamBooth/LoRA.}

The quantitative results for the DreamBooth model are shown in Tab.~\ref{tab:dreambooth_cmp}. It can be observed that the personalized effect of DreamBooth can be disrupted to some extent when noise is introduced to clean images using various methods. Among these methods, the images protected by Anti-Diffusion are visually closer to the original images, which can be seen from the highest PSNR metric. In addition, Anti-Diffusion achieves the best defense performance against DreamBooth. It causes DreamBooth to generate more meaningless images (the highest FDFR value and the lowest SER-FQA values) and disrupts DreamBooth's ability to learn the image's ID (the lowest ISM value). Additionally, DreamBooth, when trained with images protected by Anti-Diffusion, tends to generate images of the lowest quality (the highest BRISQUE, FID, and NIQE values). In summary, for face IDs on VGGFace2 and CelebA-HQ, Anti-Diffusion provides superior defense performance. 
The qualitative results in Fig.~\ref{fig:dreambooth_cmp} further support this conclusion. While methods like Photo Guard, MIST, PID, and Anti-DB offer some level of protection by reducing the visual quality of the generated images, Anti-Diffusion significantly degrades the image quality generated by the disrupted DreamBooth model and also disturbs their identities. 
As shown in Tab.~\ref{tab:lora_cmp}, we also present the quantitative defense results of different methods for LoRA. Anti-Diffusion achieves the best results on all metrics. This effectively demonstrates the good generalization ability of Anti-Diffusion against different tuning methods.

\subsubsection{Comparison on MasaCtrl/DiffEdit.}

\begin{table}[ht]
\centering
\begin{tabular}{c|c|cc|cc}
\hline
\multirow{2}{*}{Method} & \multirow{2}{*}{PSNR↑} & \multicolumn{2}{c|}{MasaCtrl}   & \multicolumn{2}{c}{DiffEdit}    \\ \cline{3-6} 
                        &                        & BRI↑       & CLI↓    & BRI↑       & CLI↓    \\ \hline
no defnese              & -                      & 22.18          & 27.44          & 16.55          & 27.65          \\ \hline
Photo             & 35.57                  & 20.40          & 27.41          & 18.76          & 26.55          \\
MIST                    & 34.87                  & 21.11          & 27.38          & 21.77          & 26.45          \\
PID                     & 35.37                  & 22.67          & 27.73          & 23.62          & 26.47          \\
Anti-DB         & 33.44                  & 25.72          & 27.42          & 24.61          & 26.69          \\
Anti-DF          &  \textbf{36.73}                  & \textbf{25.82} & \textbf{26.44} & \textbf{25.26} & \textbf{25.25} \\ \hline
\end{tabular}
\caption{Comparing the defense performance against MasaCtrl and DiffEdit on the Defense-Edit dataset. ``Photo'' and ``Anti-DF'' denotes Photo Guard and Anti-Diffusion. ``BRI'' and ``CLI'' are BRISQUE and CLIP Score.}
\label{tab:edit_cmp}
\end{table}

\begin{figure*}[ht]
\centering

\includegraphics[width=0.8\linewidth]{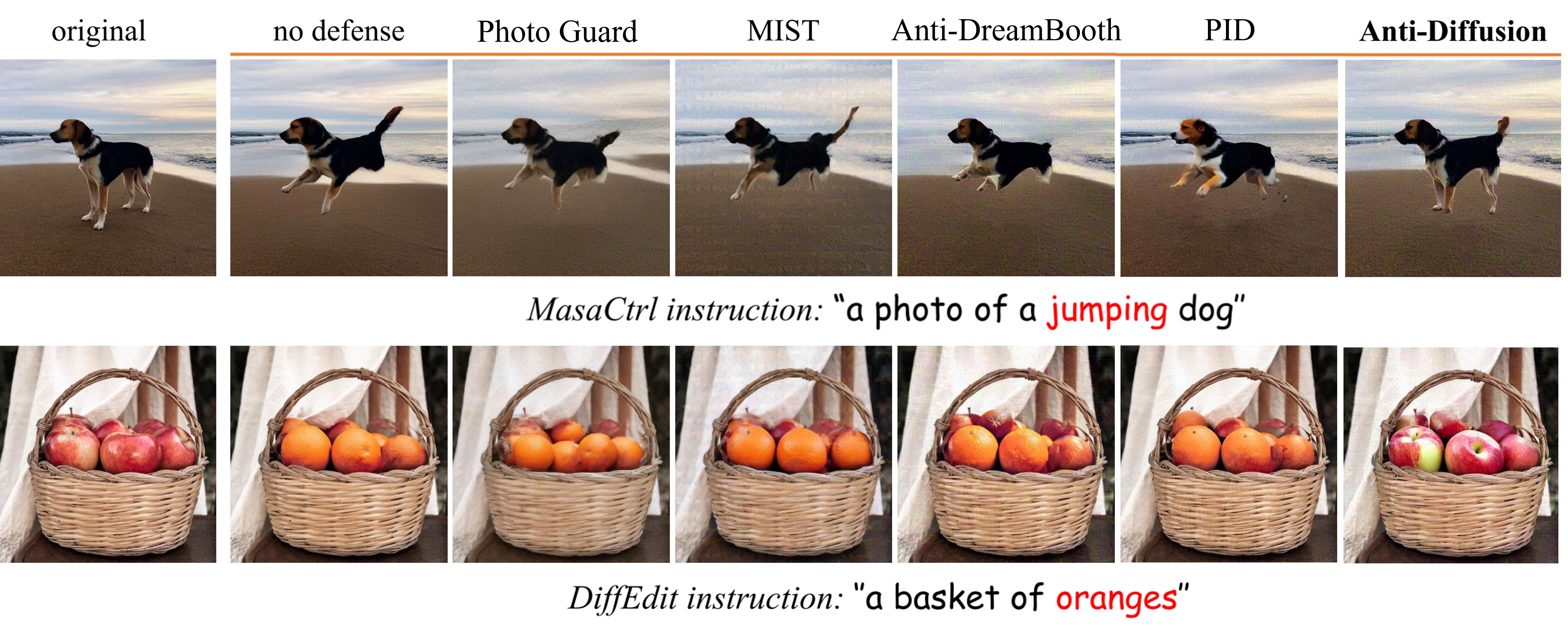}

\caption{Qualitative defense results of different defense methods on MasaCtrl and DiffEdit. The instance is from our proposed dataset Defense-Edit.
}
\label{fig:edit_cmp} 
\end{figure*}

We also compare the defense performance of different methods on MasaCtrl and DiffEdit. The quantitative results are shown in Tab.~\ref{tab:edit_cmp}, where Anti-Diffusion achieves the best performance on all three metrics. 
Specifically, Anti-Diffusion has the lowest value on the CLIP Score, indicating that when the images are protected by Anti-Diffusion, neither MasaCtrl nor DiffEdit can modify them according to the instructions. This is further validated by the qualitative results in Fig.~\ref{fig:edit_cmp}. Specifically, for the image ``dog'', when not added with noise, MasaCtrl can successfully change it from a standing posture to a jumping posture. For the protected images obtained from Photo Guard, MIST, PID, and Anti-dreamBooth, MasaCtrl can still successfully edit them. Only the images protected by Anti-Diffusion can effectively prevent MasaCtrl from editing. The same phenomenon is observed with DiffEdit, where Anti-Diffusion can effectively prevent DiffEdit from changing ``apples'' in the image to ``oranges''.

\subsection{Ablation Studies}

\begin{table}[ht]
\begin{tabular}{lc|cccc}
\hline
PT & $\mathcal{L}_{\mathrm{SDL}}$ & FDFR↑ & ISM↓ & BRISQUE↑ & FID↑   \\ \hline
   &      & 0.50  & 0.22 & 37.63    & 432.25 \\
\checkmark &      & 0.52  & 0.19 & 40.34    & 441.43 \\
   & \checkmark    & 0.53  & 0.22 & 38.45    & 432.53 \\
\checkmark  & \checkmark    & \textbf{0.62}  & \textbf{0.15} & \textbf{40.46}    & \textbf{457.13} \\ \hline
\end{tabular}
\caption{Comparing the defense performance on DreamBooth with or without PT and $\mathcal{L}_{\mathrm{SDL}}$.}
\label{tab:ablation}

\end{table}

\begin{table}[ht]
\centering
\begin{tabular}{c|cccc}
\hline
         & FDFR↑      & ISM↓          & BRISQUE↑         & FID↑         \\ \hline
Zero     & \textbf{0.62} & \textbf{0.15} & \textbf{40.46} & \textbf{457.13} \\
Noise   & 0.58          & 0.17          & 38.92          & 412.56          \\
Diagonal & 0.59          & \underline{0.15}          & 39.44          & 424.19          \\ \hline

\end{tabular}
\caption{Comparing the defense performance on different attention targets. Here, ``Noise'' means a random noise map as a target attention map, and ``Diagonal'' means a diagonal matrix where its diagonal values are set to one.}
\label{tab:target}
\end{table}

To validate the effectiveness of the PT and the SDL, we conduct comparative experiments based on DreamBooth. The details are presented in Tab.~\ref{tab:ablation}. The first experiment is a baseline experiment with a fixed prompt (i.e., ``a photo of a person''), which does not incorporate the PT and $\mathcal{L}_{\mathrm{SDL}}$. In the second row, we replace the fixed prompt with PT. For the third row, we add $\mathcal{L}_{\mathrm{SDL}}$ based on the first row. The fourth row is the final Anti-Diffusion equipped with both PT and $\mathcal{L}_{\mathrm{SDL}}$. The quantitative results of these experiments reveal that PT and $\mathcal{L}_{\mathrm{SDL}}$ play complementary roles in enhancing the defense performance.

In the experiment, we used a zero map as the target Attention map for $\mathcal{L}_{\mathrm{SDL}}$. Since cross-attention represents semantic similarity, zero-attention maps result in semantic dissimilarity between perturbed and original images. We also explored the use of random or diagonal matrices as targets. From Tab.~\ref{tab:target}, they are not as effective as zero attention maps in defense performance.

\subsection{Unexpected Scenarios}

In practical scenarios, the specific utilization of the SD models by malicious users is unpredictable. 
Therefore, in this section, we assess the defense capabilities of Anti-Diffusion in various unexpected scenarios. More results of unexpected scenarios can be found in the supplementary materials.

\subsubsection{Unexpected Version}
To evaluate the robustness of Anti-Diffusion across diverse versions of SD, we apply it to the VGGFace2 dataset using various versions of SD models, including v2.1 and v1.5. As shown in Tab.~\ref{tab:version}, Anti-Diffusion can provide sufficient protection even when the versions of SD models do not match. 

\begin{table}[ht]
\centering
\begin{tabular}{c|c|cccc}
\hline
Def.                                                              & Test & FDFR↑ & ISM↓ & BRISQUE↑ & FID↑   \\ \hline
\multirow{2}{*}{v2.1}                                                 & v2.1 & 0.62  & 0.15 & 40.46     & 457.13 \\
                                                                      & v1.5 & 0.89  & 0.03 & 43.24     & 489.45 \\ \hline
\multirow{2}{*}{v1.5}                                                 & v2.1 & 0.61  & 0.16 & 36.45     & 442.23 \\
                                                                      & v1.5 & 0.82  & 0.04 & 37.24     & 486.56 \\ \hline
\multirow{2}{*}{\begin{tabular}[c]{@{}c@{}}no\end{tabular}} & v2.1 & 0.10   & 0.66 & 17.43     & 144.02 \\
                                                                      & v1.5 & 0.06  & 0.45 & 21.43     & 134.76 \\ \hline
\end{tabular}
\caption{Comparing the defense performance on different versions of SD. The terms ``Def.'' and ``Test'' refer to the SD version for defending with Anti-Diffusion and training DreamBooth by malicious users.}
\label{tab:version}
\end{table}



\subsubsection{Unexpected Prompts}
For DreamBooth, different prompts can be used to generate various content. As illustrated in Tab.~\ref{tab:uprompt}, we introduce three additional prompts p1, p2 and p3 that are ``a photo of sks person with sad face'', ``facial close up of sks person'' and ``a photo of sks person yawning in a speech'' to evaluate the performance. We can see that Anti-Diffusion can also provide defense from different prompts in various scenarios.

\begin{table}[ht]
\centering
\begin{tabular}{c|c|cccc}
\hline
P              & \multicolumn{1}{c|}{Def.} & FDFR↑         & ISM↓          & BRISQUE↑       & FID↑            \\ \hline
\multirow{2}{*}{p1} & yes                          & \textbf{0.53} & \textbf{0.18} & \textbf{39.40} & \textbf{457.27} \\
                    & no                           & 0.09          & 0.56          & 16.34          & 169.35          \\ \hline
\multirow{2}{*}{p2} & yes                          & \textbf{0.81} & \textbf{0.08} & \textbf{27.22} & \textbf{346.21} \\
                    & no                           & 0.05          & 0.42          & 15.67          & 145.76          \\ \hline
\multirow{2}{*}{p3} & yes                          & \textbf{0.63} & \textbf{0.05} & \textbf{37.81} & \textbf{440.53} \\
                    & no                           & 0.02          & 0.31          & 18.35          & 189.21          \\ \hline
\end{tabular}
\caption{Comparing the defense performance on different prompts. ``P'' and ``Def.'' refer to prompt and defense.}
\label{tab:uprompt}
\end{table}

\section{Conclusion}

In conclusion, this paper presents Anti-Diffusion, a defense system designed to prevent images from the abuse of both tuning-based and editing-based methods. 
During the generation of the protected images, we incorporate the PT strategy to enhance defense performance, eliminating the need for manually defined prompts.
Additionally, we introduce the SDL to disrupt the semantic information of the protected images, enhancing the performance of defense against both tuning-based and editing-based methods. 
We also introduce the Defense-Edit dataset to evaluate the defense performance of current defense methods against diffusion-based editing methods. 
Through a broad range of experiments, it has been shown that  Anti-Diffusion excels in defense performance when dealing with various diffusion-based techniques in different scenarios.

\section{Acknowledgments}
This work was supported in part by Macau Science and Technology Development Fund under SKLIOTSC-2021-2023, 0022/2022/A1, and 0014/2022/AFJ; in part by Research Committee at University of Macau under MYRG-GRG2023-00058-FST-UMDF and MYRG2022-00152-FST; in part by the Guangdong Basic and Applied Basic Research Foundation under Grant 2024A1515012536.

\bibliography{aaai25}

\clearpage


\section*{Supplementary material}

# Supplementary Materials for Anti-Diffusion: Preventing Abuse of Modifications of Diffusion-Based Models

In this supplementary material for Anti-Diffusion, We firstly supplement related works and detailed training configurations of Anti-Diffusion, DreamBooth and LoRA. Then we provide a comprehensive introduction to our constructed dataset, Defense-Edit. Subsequently, we conduct a series of experiments to present both quantitative and qualitative analyses. These experiments encompass the performance evaluation on LoRA(Hu et al. 2021), the exploration of various Stable Diffusion (SD)(Rombach et al. 2022) versions in Anti-Diffusion, and the examination of different terms, prompts, and SD versions for DreamBooth(Ruiz et al. 2023). Finally, we offer additional qualitative analyses to illustrate the efficacy of our method across diverse editing tasks.

## Related Works

### Diffusion Models

Text-to-image generation models (Li et al. 2023; Ramesh et al. 2021; Gafni et al. 2022; Ding et al. 2021) aim to synthesize realistic images from natural language descriptions. Recently, Diffusion models have emerged as a promising alternative to GANs (Dhariwal and Nichol 2021; Goodfellow et al. 2014), which can generate high-quality images by reversing a stochastic Diffusion process. ControlNet(Zhang, Rao, and Agrawala 2023) is a method used for image generation that controls Diffusion models by adding additional conditions, such as pose, edge detection, depth maps, etc. It enhances the control ability of the SD model by creating "frozen" and "trainable" copies. Similar to ControlNet, T2I Adapter (Mou et al. 2023) also aligns internal knowledge with external control signals by learning lightweight adapters, thereby enhancing the controllability of the model. It does not affect the original network topology and generation capability, and has efficient operating characteristics. GLIGEN (Li et al. 2023) models the relationship between text descriptions, spatial locations, and images by introducing additional trainable self-attentions layers (Vaswani et al. 2017) on pretrained SD models. Through this method, GLIGEN can accurately control image generation based on given text descriptions and positional information. Meanwhile, a number of personalized customization generation methods based on SD have emerged, which can make the generated images have certain characteristics through a small number of samples. Textual Inversion can achieve personalized generation over generated images by learning the representation of new "words" in the embedding space of the text encoder. It only needs to use 3-5 related images provided by the user to guide the generation by learning these concepts. DreamBooth can fine tune the SD model with a small number of graphs to introduce the desired concepts into a rare used word. Similarly, with only a small number of samples required, LoRA accelerates the training process of the model and significantly reduces the storage cost of model files by slightly modifying and recalculating the weight layers of the original base model.

In addition to generating images, the editing method powered by SD can also quickly edit input images through instructions, such as the actions of characters or the number of items in the image. MasaCtrl is a method that does not require fine-tuning and can achieve consistent image generation and editing simultaneously. It converts Self-Attention in the Diffusion model into mutual Attention to query relevant local content and textures in the source image, achieving consistency between the edited image and the original image. DiffEdit can replace the target in the image according to different text descriptions. It can automatically generate a mask that highlights the input image area that needs to be edited, and then use this mask and a text guided Diffusion model to achieve semantic image editing.

In these works, DreamBooth and LoRA is widely used due to its excellent generation ability and few-shot learning ability. Moreover, image editing methods cannot be ignored since MasaCtrl can immediately edit images without additional tuning. Therefore, we focus on the protecting against the malicious use of these methods.

### Defense Against SD

While SD brings efficient personalized generation and editing capabilities, defense against SD is also being studied to protect images from unauthorized learning. Photo Guard (Salman et al. 2023) generates adversarial noise for the encoder and denoising process in SD. Similar to Photo Guard, MIST (Liang et al. 2023) interferes with both processes. At the same time, MIST constructs a specific watermark that not only interferes with the image generation process of SD, but also marks specific watermarks on the generated image.

Yet, for DreamBooth, its training process constantly updates the parameters of the SD, which makes the above two methods prone to failure. On the basis of interfering with the denoising process of SD, Anti-DreamBooth (Van Le et al. 2023) simulates the training of DreamBooth through the constructed dataset, dynamically updating the parameters of SD. Through this alternating defense and training, better defense performance have been achieved against DreamBooth. PID (Li et al. 2024) generates adversarial noise to disturb VAE encoder of SD, which has the advantage of ignoring the impact of prompts and providing broader protection than Anti-DB. However, PID solely relies on VAE and does not participate in the diffusion process. Its defense performance is greatly affected by VAE versions. Given the unpredictability of both prompts and model versions in practical applications, a more comprehensive defense strategy is necessary. At the same time, these methods mainly focus on tuning models and overlook diffusion-based editing methods, emphasizing the necessity for a more extensive defense approach against both threats. These methods demonstrate their ability to protect images from unauthorized learning and editing, but there are still some shortcomings. PhotoGuard and Mist perform adversarial examples generation on fixed parameter of SD, which results in poor protection on methods such as DreamBooth that require tuning, while Anti DreamBooth requires users to construct additional datasets, which is not very practical for practical applications. And Anti-DB only interferes with the denoising process of SD with manually selected prompt, without interfering with the semantic features of adversarial examples, which makes its privacy protection ability and resistance to editing methods not outstanding enough.

## Training Configurations

During the Anti-Diffusion training process, it takes a total of 5 epochs to obtain the final perturbed images. Each epoch consists of 30 iterations for Prompt Tuning, 10 iterations for PGD in adversarial noise optimization, and 20 iterations for UNet update. For the PGD part, we utilize a value of $\alpha = 0.002$ and a default noise budget of $\eta = 0.05$. The adversarial noise $\delta(i)$ is optimized using the Projected Gradient Descent (PGD) scheme, with 10 iterations of PGD employed. Our method utilizes a value of $\alpha = 0.002$ and a default noise budget of $\eta = 0.05$ for the PGD part. There are 30 iterations for Prompt Tuning, 20 iterations for UNet tuning at each epoch and it takes 5 epochs for the protection. The whole training process, when executed on an NVIDIA A6000 GPU, takes about 3 minutes to complete. During the optimization of the Prompt Tuning strategy, the dimension of the optimized token embedding is $77 \times 1024$. For the Semantic Disturbance Loss, we only extract Cross-Attention maps at a scale of $16 \times 16$ to save memory.

DreamBooth is trained by utilizing a batch size of 2 and a learning rate of $5 \times 10^{-7}$ across $1,000$ training steps. The most recent version of SD (v2.1) is employed as the pre-trained generator by default. Unless otherwise indicated, "a photo of sks person" and "a photo of person" are set as the training instance prompt and prior prompt, respectively. For the training of LoRA, it takes 400 epochs with a learning rate of $1e^{-5}$. The settings for the training instance prompt and prior prompt are the same as DreamBooth.

## Details of the Dataset Defense-Edit

To evaluate the performance of defense methods on SD-based editing, we collect a dataset named Defense-Edit with 50 pairs of images and prompts. In order to simulate different scenarios, we collect 30 images from real photos, and 20 images generated using SD. For real photos, we select them from CelebA-HQ(Karras et al. 2017), VGGFace2(Cao et al. 2018), and TEdBench(Kawar et al. 2023) to cover categories such as faces, animals (e.g., dogs and horses), and objects (e.g., apples and boxes). For the generated images, we adopt SD v2.1 to generate them with our constructed prompts. In the process of creating instructions, to have an extensive coverage of a wide range of conceivable scenarios, we incorporate transformations that include changes in facial expressions, item substitutions, additions, deletions, and pose adjustments. To avoid situations where the editing effect itself is not ideal, we select data that can be well edited on MasaCtrl(Cao et al. 2023) or DiffEdit(Couairon et al. 2023). Some representative images with their instructions from our Defense-Edit dataset and their edited versions using DiffEdit and MasaCtrl can be found in Fig. 1.

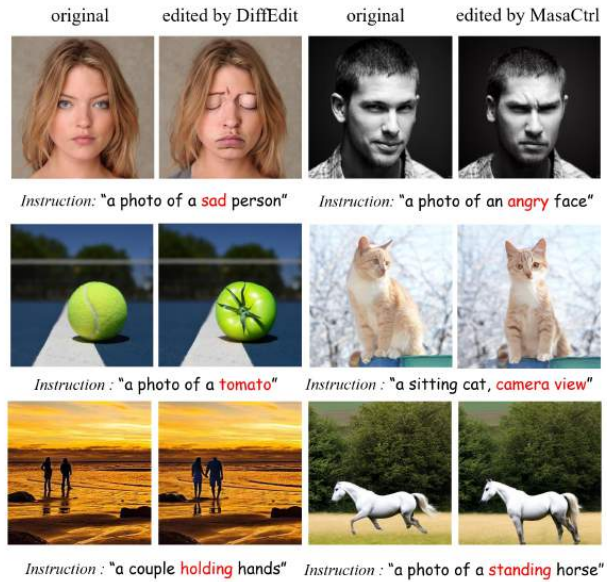

Figure 1: Examples of Defense-Edit with original images, instructions, and editing effects.

## Additional quantitative results

In the main paper, we analyze the performance of Anti-Diffusion, utilizing the most recent SD version, v2.1. The term "sks" is employed during the training phase of the DreamBooth, and the prompt "a photo of sks person" is used during the inference stage. In this section, we provide more defense results on LoRA. Then we explore various terms,

prompts, and versions of SD on DreamBooth for evaluating the performance of Anti-Diffusion.

### Defense performance on LoRA

In addition to applying LoRA on VGGFace2, we further investigate the effectiveness of different defense methods by applying LoRA on the CelebA-HQ dataset. Other baselines like Anti-DreamBooth, PID, Photo Guard and MIST are also adopted as an comparison. As demonstrated in Tab. 1, Anti-Diffusion makes LoRA to generate more meaningless images (the highest FDFR(Deng et al. 2020) value), and disrupts LoRA's ability to learn the image's ID (the lowest ISM(Deng et al. 2019) and SER-FQA(Terhorst et al. 2020) values). In addition, the LoRA, when trained with images perturbed using Anti-Diffusion, tends to generate images of the lowest quality (the highest BRISQUE(Mittal, Moorthy, and Bovik 2012), FID(Heusel et al. 2017) and NIQE(Mittal, Soundararajan, and Bovik 2012) values). This suggests that Anti-Diffusion has good generalizability when applied to another diffusion-based method.

### Performance across different terms for DreamBooth

In the process of training DreamBooth, a term(e.g.,sks) is required to encapsulate the semantic information of the target object. The selection of different terms may influence the generative capacity of the model. To ensure a comprehensive analysis of robustness on different terms and validate the efficacy of our method, we experiment with a broader range of terms. Specifically, we evaluate the effectiveness of our method on DreamBooth using diverse terms such as "sks", "t@t", and "dbz" on both CelebA-HQ and VGGFace2 datasets. Anti-DreamBooth, which has the best performance among the baselines, is selected for the comparison. The quantitative results are shown in Tab. 2 and Tab. 3. With the defense of Anti-Diffusion, DreamBooth generates more nonsensical images, as indicated by the highest FDFR value on both datasets CelebA-HQ and VGGFace2,. Also it limits DreamBooth's capacity to learn the image's ID, as evidenced by the lowest ISM value. Furthermore, DreamBooth trained with images protected by Anti-Diffusion tends to generate lower quality images, as shown by the highest BRISQUE, FID, and NIQE values. Overall, Anti-Diffusion demonstrates superior defense performance against DreamBooth under various terms, on both CelebA-HQ and VGGFace2 datasets.

### Performance across different prompts for DreamBooth

For DreamBooth, different prompts can be used to generate different content. As illustrated in Tab. 4, we introduce two additional prompts: "a photo of sks person with sad face" and "a photo of sks person with angry face" to assess the performance. When compared to Anti-DreamBooth, Anti-Diffusion achieves superior defense performance in these varied prompt scenarios.

### Performance across different SD versions

In real-world situations, it's impractical to know the specific version of SD that will be employed for training DreamBooth. To evaluate the robustness of our defense method across diverse versions of SD, we apply Anti-Diffusion on the VGGFace2 dataset across various versions of SD models, including v2.1, v1.5, and v1.4. The defense effects of images protected with different versions of SD are listed in Tab. 5, Tab. 6, and Tab. 7 respectively. Compared to images with no defense, Anti-Diffusion can prove effective across various SD versions. It remains effective when the SD version for defense differs from that of DreamBooth. Notably, our method outperforms Anti-DreamBooth on all the metrics, especially on ISM, suggesting that DreamBooth is more difficult to imitate the facial features from images protected by Anti-Diffusion.

| Method | PSNR↑ | FDFR↑ | ISM↓ | SER-FQA↓ | Brisques↑ | FID↑ | NIQE↑ |
|---|---|---|---|---|---|---|---|
| no defense | — | 0.02 | 0.67 | 0.74 | 17.14 | 142.67 | 4.22 |
| photo guard | 34.24 | 0.04 | 0.62 | 0.65 | 17.71 | 143.58 | 5.48 |
| MIST | 35.31 | 0.03 | 0.53 | 0.54 | 29.96 | 239.86 | 4.53 |
| PID | 34.72 | 0.18 | 0.51 | 0.55 | 36.42 | 243.58 | 6.29 |
| Anti-DreamBooth | 35.76 | 0.10 | 0.47 | 0.49 | 38.67 | 257.73 | 6.68 |
| Anti-Diffusion | **36.76** | **0.14** | **0.43** | **0.43** | **42.27** | **275.59** | **7.11** |

Table 1: Defense performance comparison when applying LoRA on CelebA-HQ.

| Terms | Method | FDFR↑ | ISM↓ | SER-FQA↓ | Brisques↑ | FID↑ | NIQE↑ |
|---|---|---|---|---|---|---|---|
| t@t | Anti-Diffusion | **0.46** | **0.23** | **0.31** | **40.66** | **388.58** | **5.55** |
|  | Anti-DreamBooth | 0.40 | 0.26 | 0.49 | 36.96 | 358.93 | 5.37 |
|  | no defense | 0.10 | 0.68 | 0.72 | 15.37 | 139.33 | 4.24 |
| dbz | Anti-Diffusion | **0.62** | **0.14** | **0.20** | **40.35** | **430.70** | **5.30** |
|  | Anti-DreamBooth | 0.57 | 0.23 | 0.34 | 37.32 | 413.54 | 4.98 |
|  | no defense | 0.11 | 0.69 | 0.74 | 16.47 | 142.57 | 4.46 |
| sks | Anti-Diffusion | **0.62** | **0.15** | **0.18** | **40.46** | **457.13** | **5.27** |
|  | Anti-DreamBooth | 0.60 | 0.24 | 0.31 | 37.41 | 436.34 | 5.05 |
|  | no defense | 0.10 | 0.66 | 0.73 | 17.43 | 144.02 | 4.12 |

Table 2: Defense performance comparison across different terms for DreamBooth on VGGFace2.

| Terms | Method | FDFR↑ | ISM↓ | SER-FQA↓ | Brisques↑ | FID↑ | NIQE↑ |
|---|---|---|---|---|---|---|---|
| t@t | Anti-Diffusion | **0.22** | **0.43** | **0.57** | **39.65** | **280.38** | **5.98** |
|  | Anti-DreamBooth | 0.17 | 0.56 | 0.64 | 37.64 | 263.76 | 5.51 |
|  | no defense | 0.04 | 0.67 | 0.73 | 19.29 | 155.84 | 4.53 |
| dbz | Anti-Diffusion | **0.29** | **0.25** | **0.50** | **41.25** | **332.59** | **5.99** |
|  | Anti-DreamBooth | 0.21 | 0.44 | 0.54 | 38.08 | 316.04 | 5.49 |
|  | no defense | 0.13 | 0.70 | 0.61 | 21.22 | 139.56 | 4.52 |
| sks | Anti-Diffusion | **0.58** | **0.26** | **0.38** | **40.93** | **352.83** | **5.96** |
|  | Anti-DreamBooth | 0.54 | 0.41 | 0.29 | 38.34 | 336.12 | 5.56 |
|  | no defense | 0.07 | 0.63 | 0.73 | 17.00 | 147.82 | 4.72 |

Table 3: Defense performance comparison across different terms for DreamBooth on CelebA-HQ.

| Method | PSNR↑ | FDFR↑ | ISM↓ | SER-FQA↓ | BRISQUE↑ | FID↑ | NIQE↑ |
|---|---|---|---|---|---|---|---|
| | "a photo of sks person with sad face" | | | | | | |
| Anti-Diffusion | **35.91** | **0.53** | **0.18** | **0.43** | **39.40** | **457.27** | **5.13** |
| Anti-DreamBooth | 34.55 | 0.47 | 0.23 | 0.45 | 36.32 | 425.75 | 5.01 |
| no defense | - | 0.09 | 0.56 | 0.71 | 16.34 | 169.35 | 4.11 |
| | "a photo of sks person with angry face " | | | | | | |
| Anti-Diffusion | **35.91** | **0.55** | **0.16** | **0.41** | **39.51** | **432.54** | **5.16** |
| Anti-DreamBooth | 34.55 | 0.49 | 0.17 | 0.50 | 37.42 | 412.93 | 5.12 |
| no defense | - | 0.11 | 0.59 | 0.75 | 15.36 | 173.69 | 4.08 |

Table 4: Defense performance comparison across different prompts for DreamBooth.

| SD | Method | PSNR↑ | FDFR↑ | ISM↓ | SER-FQA↓ | BRISQUE↑ | FID↑ | NIQE↑ |
|---|---|---|---|---|---|---|---|---|
| v2.1 | Anti-Diffusion | **35.91** | **0.62** | **0.15** | **0.18** | **40.46** | **457.13** | **5.27** |
| | Anti-DreamBooth | 34.55 | 0.60 | 0.24 | 0.31 | 37.41 | 436.34 | 5.05 |
| v1.5 | Anti-Diffusion | **36.35** | **0.61** | **0.16** | **0.21** | **36.45** | **442.23** | **5.01** |
| | Anti-DreamBooth | 34.59 | 0.57 | 0.28 | 0.34 | 35.19 | 425.27 | 4.98 |
| v1.4 | Anti-Diffusion | **35.90** | **0.60** | **0.16** | **0.21** | **35.68** | **431.64** | **5.23** |
| | Anti-DreamBooth | 34.57 | 0.54 | 0.29 | 0.33 | 35.24 | 412.76 | 5.01 |
| - | no defense | - | 0.10 | 0.66 | 0.73 | 17.43 | 144.02 | 4.12 |

Table 5: Defense performance comparison on SD v2.1 based DreamBooth.

| SD | Method | PSNR↑ | FDFR↑ | ISM↓ | SER-FQA↓ | BRISQUE↑ | FID↑ | NIQE↑ |
|---|---|---|---|---|---|---|---|---|
| v2.1 | Anti-Diffusion | **35.91** | **0.89** | **0.03** | **0.11** | **43.24** | **489.45** | **5.95** |
|  | Anti-DreamBooth | 34.55 | 0.71 | 0.13 | 0.22 | 30.53 | 432.86 | 5.34 |
| v1.5 | Anti-Diffusion | **36.35** | **0.82** | **0.04** | **0.07** | **37.24** | **486.56** | **5.89** |
|  | Anti-DreamBooth | 34.59 | 0.67 | 0.18 | 0.24 | 28.72 | 401.39 | 5.11 |
| v1.4 | Anti-Diffusion | **35.90** | **0.82** | **0.05** | **0.09** | **36.76** | **473.36** | **5.73** |
|  | Anti-DreamBooth | 34.57 | 0.62 | 0.20 | 0.22 | 29.56 | 392.64 | 5.01 |
| - | no defense | - | 0.06 | 0.45 | 0.67 | 21.43 | 134.76 | 4.67 |

Table 6: Defense performance comparison on SD v1.5 based DreamBooth.

| SD | Method | PSNR↑ | FDFR↑ | ISM↓ | SER-FQA↓ | BRISQUE↑ | FID↑ | NIQE↑ |
|---|---|---|---|---|---|---|---|---|
| v2.1 | Anti-Diffusion | **35.91** | **0.94** | **0.02** | **0.09** | **41.54** | **496.74** | **6.13** |
|  | Anti-DreamBooth | 34.55 | 0.81 | 0.07 | 0.12 | 26.42 | 456.78 | 5.54 |
| v1.5 | Anti-Diffusion | **36.35** | **0.83** | **0.04** | **0.08** | **35.62** | **454.41** | **5.75** |
|  | Anti-DreamBooth | 34.59 | 0.73 | 0.16 | 0.24 | 30.35 | 435.24 | 5.38 |
| v1.4 | Anti-Diffusion | **35.90** | **0.81** | **0.05** | **0.07** | **37.45** | **489.82** | **5.91** |
|  | Anti-DreamBooth | 34.57 | 0.77 | 0.17 | 0.21 | 28.45 | 442.54 | 5.23 |
| - | no defense | - | 0.06 | 0.49 | 0.65 | 19.67 | 112.84 | 4.31 |

Table 7: Defense performance comparison on SD v1.4 based DreamBooth.

## Additional Qualitative Results

### Defense Performance on DreamBooth

As depicted in Fig. 2, we demonstrate the effectiveness of various defense methods when employing DreamBooth on the CelebA-HQ datasets. Under different character identities, with the defense of Anti-Diffusion, images generated by DreamBooth contain more noise and their facial features are more difficult to be recognized. It indicates that Anti-Diffusion achieves superior defense performance when utilizing DreamBooth.

### Defense Performance on LoRA

As depicted in Fig. 3, we demonstrate the effectiveness of various defense methods when employing LoRA on the VG-GFace2 and CelebA-HQ datasets. Under different datasets and character identities, with the defense of Anti-Diffusion, images generated by LoRA contain more noise and their facial features are more difficult to be recognized. It indicates that Anti-Diffusion achieves superior defense performance when utilizing LoRA.

### Performance across Different Terms for DreamBooth

We provide more visual results of the performance across different terms of DreamBooth. As depicted in Fig. 4, the top two rows display the results on the CelebA-HQ dataset, while the bottom two rows present the results on the VGGFace2 dataset. It can be seen that the defense of Anti-Diffusion is effective across different terms. Compared to Anti-DreamBooth, the defense of Anti-Diffusion forces DreamBooth to generate images with more noise and less recognizable faces.

### Performance across Different Prompts for DreamBooth

Here we show the visual results of our defense methods across different prompts for DreamBooth. As illustrated in Fig. 5, with the defense of Anti-Diffusion, DreamBooth generates images with increased perturbations. This indicates that our method can effectively defense across a variety of prompts and outperforms Anti-DreamBooth.

### Performance across Different SD Versions

We present more visual results across different SD versions. As illustrated in Fig. 6, images generated by DreamBooth with the defense of Anti-Diffusion have more notable artifacts. This shows the robustness of our defense method across different versions of the SD model.

### Performance on Editing Methods

Here we show the defense performance of Anti-Diffusion against MasaCtrl and DiffEdit on Defense-Edit dataset. As illustrated in Fig. 7, following the instructions, MasaCtrl and DiffEdit can edit the images with no defense or defensed by Anti-DreamBooth. Through the defense of Anti-Diffusion, these editing methods fail. For example, in the third row of Fig. 7, with no defense or with Anti-DreamBooth, the tennis ball is successfully transformed into to a green tomato. In contrast, with Anti-Diffusion, the tennis ball remains unchanged. This means that Anti-Diffusion has better defense performance across these editing methods.
.

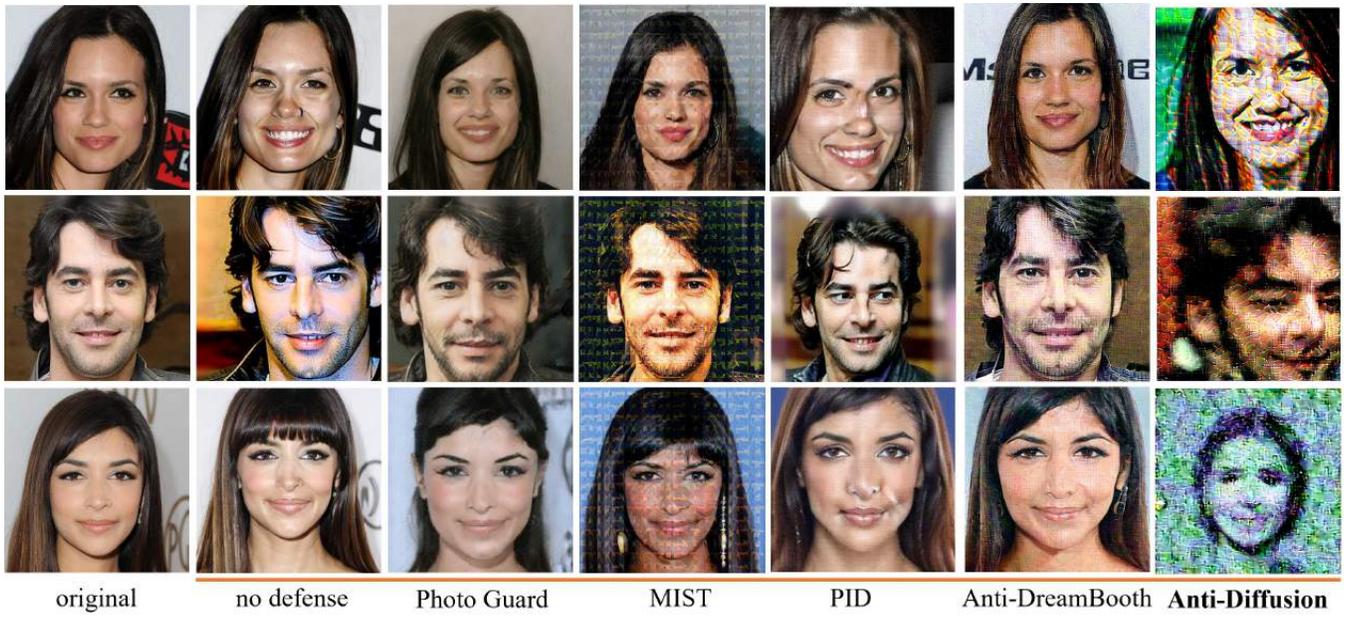

Figure 2: Defense performance comparison when applying DreamBooth on dataset VGGFace2.

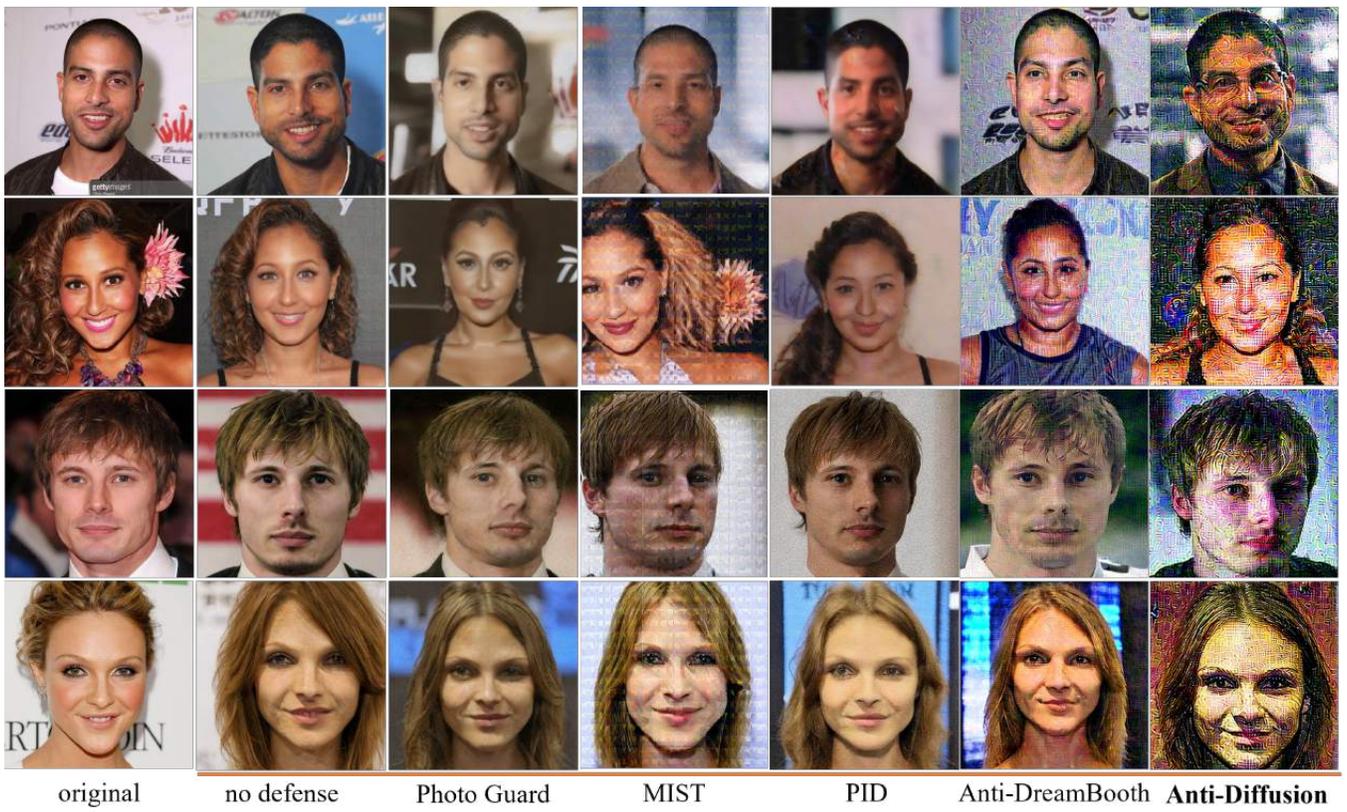

Figure 3: Defense performance comparison when applying LoRA on dataset VGGFace2 ( the first two rows) and dataset CelebA-HQ ( the last two rows).

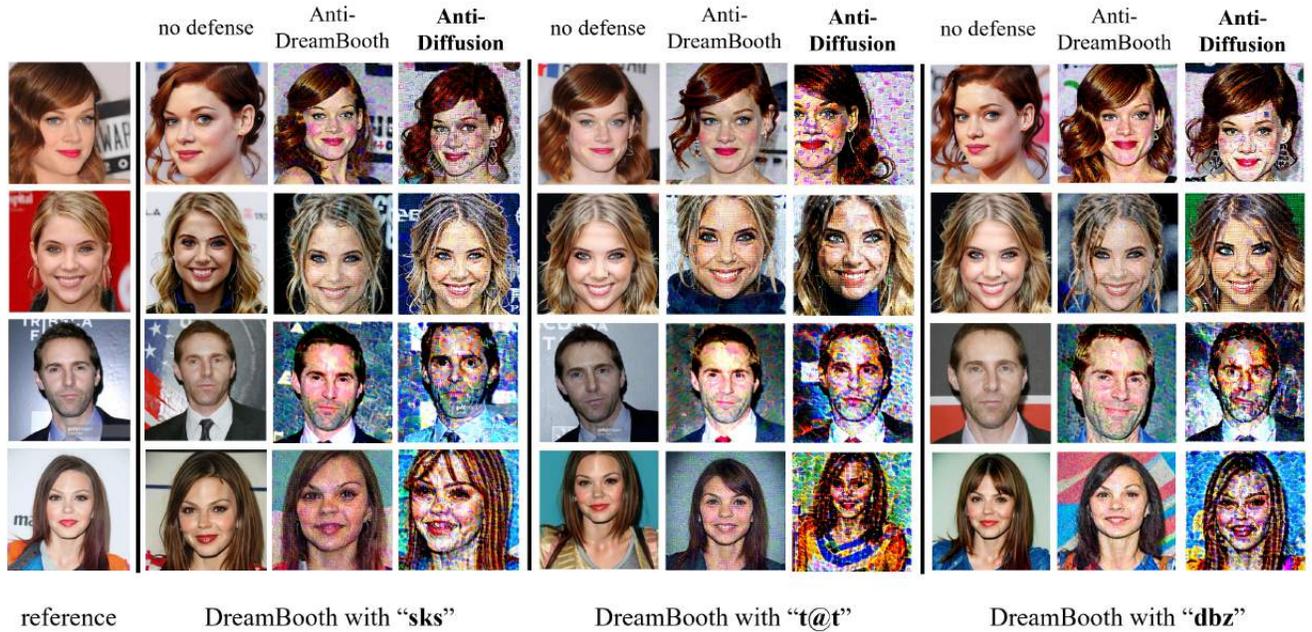

Figure 4: Defense performance comparison across different terms for DreamBooth.

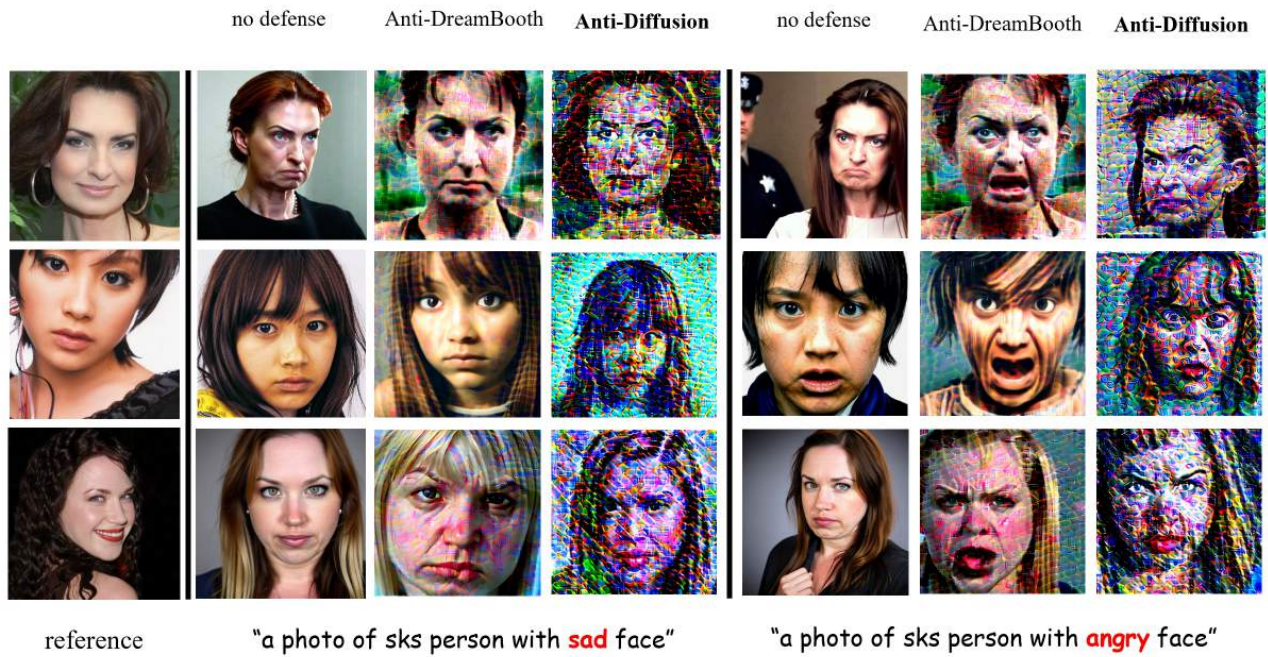

Figure 5: Defense performance comparison across different prompts for DreamBooth.

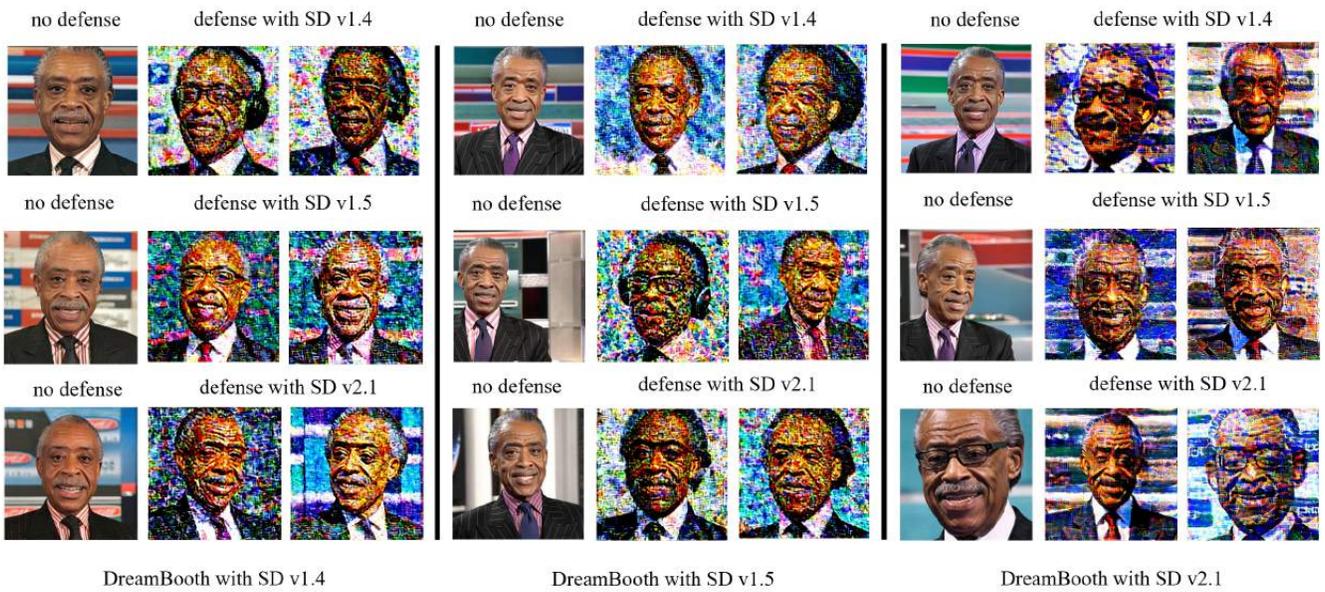

Figure 6: Defense performance comparison across different SD versions for DreamBooth.

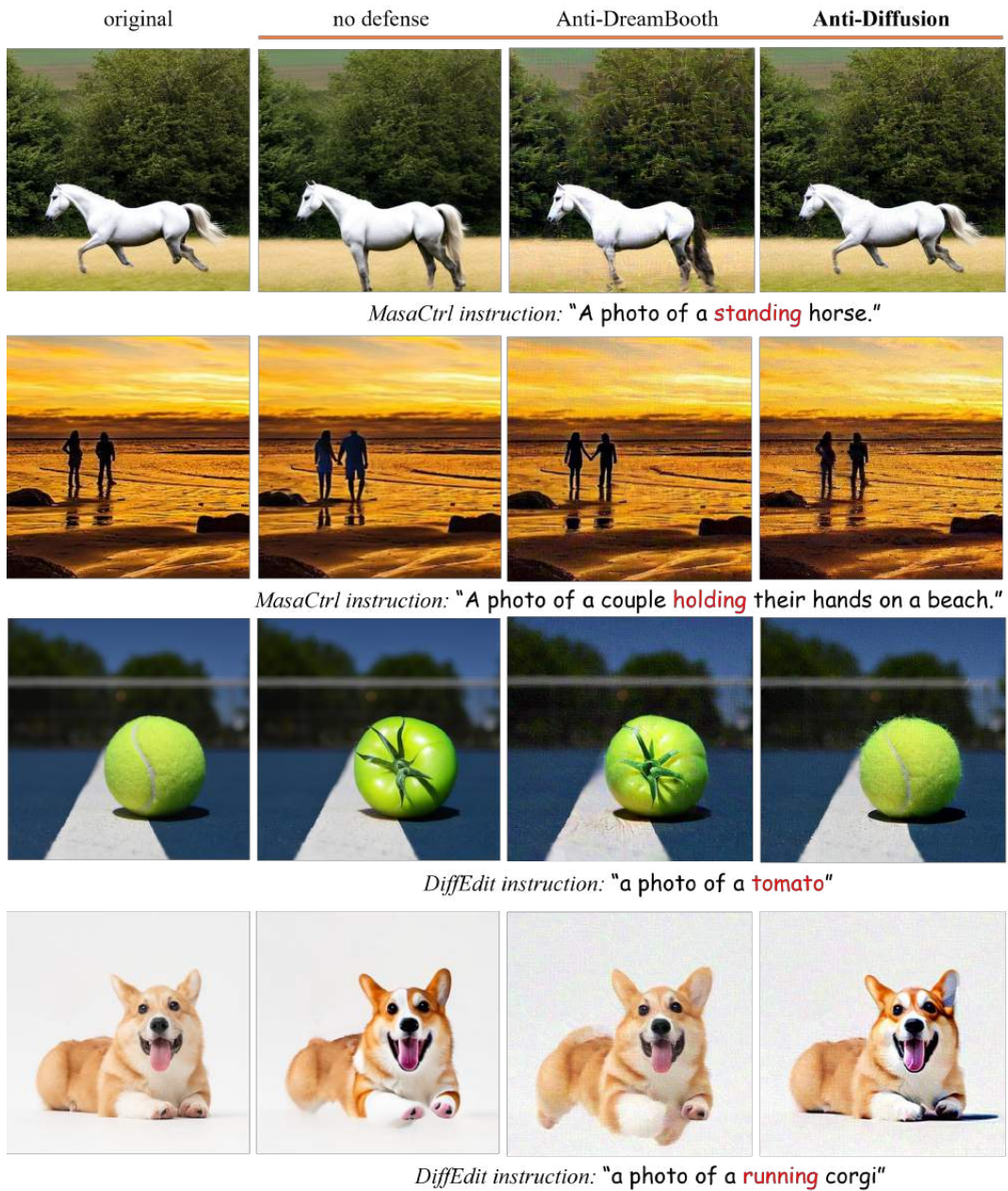

Figure 7: Qualitative results across different editing method.



\end{document}